\documentclass{article}
\usepackage{ijcai26}

\usepackage{times}
\usepackage{soul}
\usepackage{url}
\usepackage[hidelinks]{hyperref}
\usepackage[utf8]{inputenc}
\usepackage[small]{caption}
\usepackage{graphicx}
\usepackage{amsmath}
\usepackage{amsthm}
\usepackage{amssymb}
\usepackage{booktabs}
\usepackage{algorithm}
\usepackage{algorithmic}
\usepackage[switch]{lineno}
\usepackage{xspace}
\usepackage{multirow}
\usepackage{tcolorbox}

\newcommand{\methodname}{CTO\xspace}

\title{Improving Code Translation with Syntax-Guided and Semantic-aware\\ Preference Optimization}

\author{
Yuhan Wu$^1$
\and
Huan Zhang$^1$
\and
Wei Cheng$^1$
\and
Chen Shen$^1$
\and
Jingyue Yang$^1$
\And
Wei Hu$^{1,2,}$\thanks{Corresponding author}\\
\affiliations
$^1$State Key Laboratory for Novel Software Technology, Nanjing University, China\\
$^2$National Institute of Healthcare Data Science, Nanjing University, China\\
\emails
\{yhwu, zhanghuan, wchengcs, cshen, jyyang\}.nju@gmail.com, whu@nju.edu.cn
}

\begin{document}

\maketitle

\begin{abstract}
LLMs have shown immense potential for code translation, yet they often struggle to ensure both syntactic correctness and semantic consistency. 
While preference-based learning offers a promising alignment strategy, it is hindered by unreliable semantic rewards derived from sparse test cases or restrictive reference translations. 
We argue that a robust semantic reward for code translation must be derived directly from the source code. 
In this paper, we propose \methodname to improve \underline{c}ode \underline{t}ranslation with syntax-guided and semantic-aware preference \underline{o}ptimization. 
Through contrastive learning, we train a cross-lingual semantic model to directly assess functional equivalence between source and translated code. 
By formulating code translation as a multi-objective optimization problem, this robust semantic signal is seamlessly unified with compiler-based syntactic feedback within the direct preference optimization framework. 
Extensive experiments on C++, Java, and Python translations demonstrate that \methodname significantly outperforms existing baselines and alternative preference optimization strategies.

\end{abstract}

\section{Introduction}
\label{sec:intro}
Code translation, the process of migrating functionality from a source programming language to a target language, is a crucial task in modern software engineering. 
It promotes code reuse, facilitates legacy system modernization, and enables cross-language interoperability in an increasingly heterogeneous software ecosystem \cite{Nguyen2014Migrating,zhu2022xlcost,yan2023codetransocean}. 
While recent advances in pre-trained language models have shown immense potential \cite{lu2021codexglue,zhu2022multilingual,zheng2023codegeex,roziere2020transcoder}, their practical application is frequently hindered by a key challenge: ensuring that the translated code is not only syntactically correct but also semantically equivalent to the source. 
Even state-of-the-art large language models (LLMs) often make errors due to inadequate understanding of the syntactic and semantic nuances across different programming languages \cite{pan2024lost,yang24unitrans}.

\begin{figure}
\centering
\includegraphics[width=\linewidth]{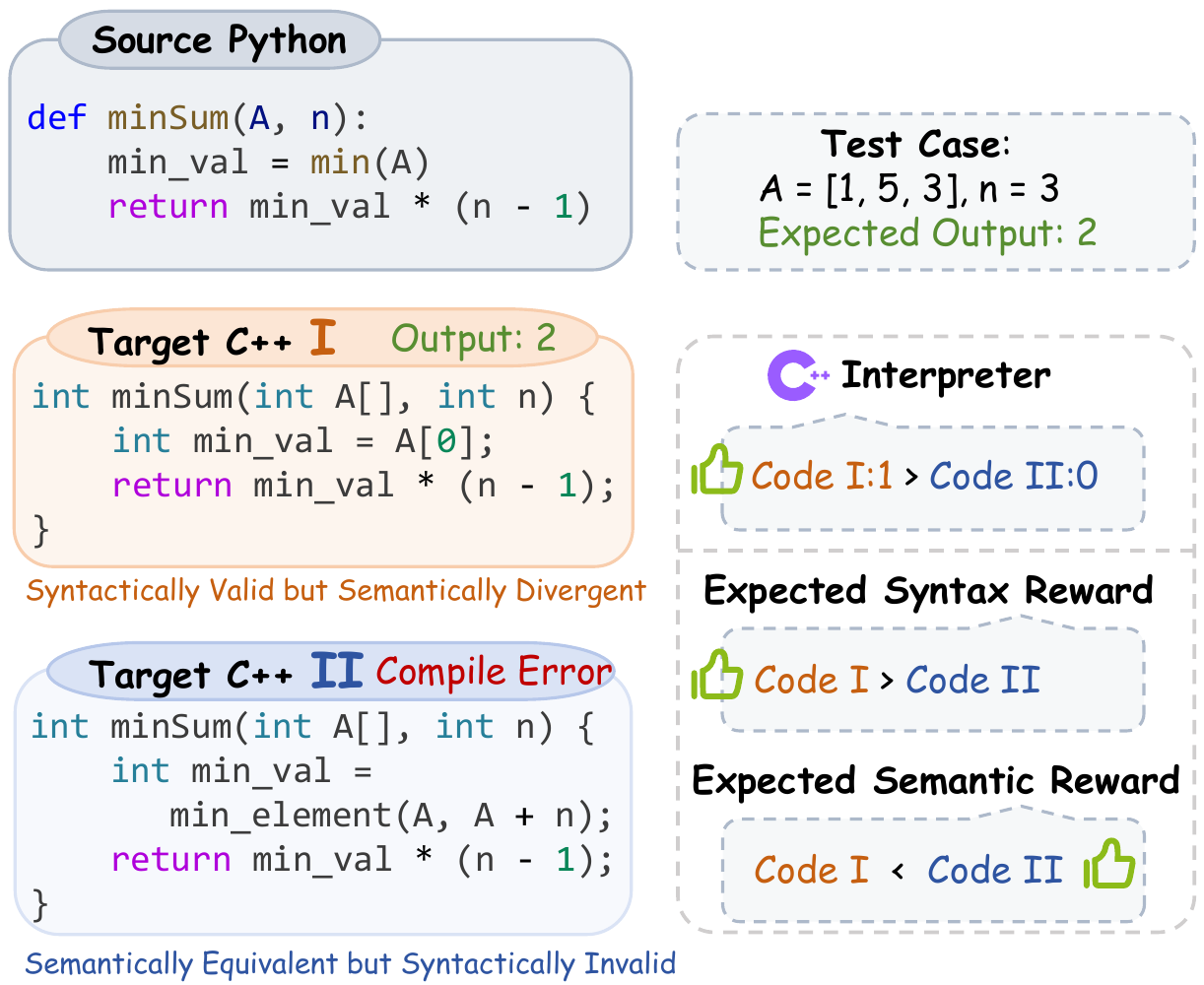}
\caption{A motivating example illustrating the entanglement of syntax and semantics in execution-based reward. \textbf{Target C++ I} is syntactically valid but semantically flawed; however, it exploits sparse test cases (where \texttt{A[0]} happens to be the minimum) to achieve a false positive pass (\textit{reward hacking}). Conversely, \textbf{Target C++ II} maintains semantic equivalence (using \texttt{min\_element}) but receives a zero reward due to a compilation error.}
\label{fig:example}
\end{figure}

To address this, a natural evolution is to move beyond standard supervised finetuning and adopt preference-based learning.
By leveraging reward signals derived from user judgments or task-specific criteria, reinforcement learning from human feedback (RLHF) \cite{ouyang2022rlhf} has been widely employed to align model outputs with human preferences.
However, applying this paradigm to code translation raises a fundamental question: \textbf{How can we define and obtain reliable reward signals for both syntactic correctness and semantic consistency?}
For syntax, compiler feedback serves as an infallible oracle, providing a deterministic and binary signal of syntactic correctness \cite{shojaee2023executionbased}.
For semantics, however, the path is fraught with challenges.

As illustrated in Figure \ref{fig:example}, prevailing strategies for semantic rewards are built on flawed foundations.
The most common strategy is to derive rewards from test cases, treating test outcomes as a proxy for functional correctness \cite{le2022coderl,leonidas2025optimise,zhang2025codedpo}. 
However, real-world test suites are often sparse and exhibit low coverage, which can lead to reward hacking \cite{Ma2025coderm}, where the model may overfit to these limited test cases without preserving the intended semantics of the source code.
In the context of code translation, relying on test cases leads to a critical entanglement of syntax and semantics. A single syntactical divergence in the target language often invalidates the entire execution, yielding a zero pass rate regardless of semantic accuracy. 
Consequently, the model receives a binary reward that fails to decouple semantic correctness from syntactic validity, making it impossible to quantify the magnitude of semantic deviation.
An alternative is to use the reference translation as a semantic anchor. 
However, this approach gives rise to two flawed strategies. 
The first strategy compares the generated code to the reference using text-based similarity metrics such as CodeBLEU \cite{Ren2020codebleu}. This strategy is fundamentally superficial, as it rewards lexical similarity over true functional equivalence. 
The second, more sophisticated strategy introduces a monolingual semantic model to evaluate the similarity between candidate translations and the reference. 
However, it implicitly assumes that the reference translation is a perfect and complete representation of the source semantics. 
This assumption introduces a \textit{semantic bottleneck} that restricts output diversity and may even propagate errors in the reference.
We argue that a robust semantic reward should be disentangled from such flawed proxies and instead be derived directly from the source code itself.
We propose \methodname, a novel approach that improves \underline{c}ode \underline{t}ranslation via syntax-guided and semantic-aware preference \underline{o}ptimization.

We reformulate code translation as a multi-objective preference optimization problem, explicitly modeling both syntax and semantics as distinct objectives. 
It leverages compiler feedback to construct syntax-aware preference pairs.
For semantic alignment, it trains a cross-lingual semantic model via contrastive learning to assess functional equivalence directly between the source and translated candidates. 
It injects the semantic reward difference of a preference pair directly into the learning process, biasing the preference strength to unify syntactic and semantic objectives within the direct preference optimization (DPO) \cite{rafailov2024direct} framework.

We conduct extensive experiments on pairwise code translation between C++, Java, and Python.
Results demonstrate that \methodname outperforms existing approaches and alternative preference optimization strategies, achieving a superior alignment with the intended functionality of the source code.

To summarize, our main contributions are as follows:
\begin{itemize}

\item We propose \methodname, a novel approach that improves code translation via syntax-guided and semantic-aware preference optimization. It introduces a reward-biasing mechanism within DPO, enabling unified optimization of both syntactic correctness and semantic consistency.

\item We train a cross-lingual semantic model as a robust reward source. It directly evaluates functional equivalence between the source code and translation candidates, overcoming the fundamental limitations of reward signals derived from other proxies.

\item Our experiments show that \methodname improves translation accuracy by up to 3.66\% on the TransCoder-Test dataset and 4.27\% on HumanEval-X with the finetuned CodeT5 model. 
With a larger finetuned CodeLlama-7B model, it yields accuracy gains of 5.60\% and 6.70\%, respectively.
\end{itemize}

\section{Related Work}
\label{sec:related_work}

\paragraph{Code Translation.}
Early works \cite{roziere2020transcoder,roziere2022leveraging} on code translation primarily focus on self-supervised learning. Building upon this, several works \cite{szafraniec2022code,huang2023program,liu2023syntax} explore incorporating code structure information to enhance code representations for unsupervised code translation. 
These methods demonstrate the feasibility of learning cross-language representations without parallel data. Despite their effectiveness, unsupervised approaches often struggle with collecting enormous amounts of code corpora and high computational resource consumption. 

With the availability of high-quality parallel datasets \cite{zhu2022xlcost,yan2023codetransocean,mohammad2024xcodeeval,weixiang2024codescope} and pre-trained code models \cite{wang2021codet5,guo2021graphcodebert,zheng2023codegeex,feng2020codebert}, supervised finetuning has become a paradigm for code translation. 
Building on this foundation, recent studies explore incorporating reinforcement learning (RL) techniques \cite{shojaee2023executionbased} and variational inference techniques \cite{du2024joint} to further enhance the performance. 
Besides, a few recent works \cite{ibrahimzada2024repository,wang2024repotransbench} focus on repository-level code translation, aiming to facilitate complex codebase migration.
Our work follows the supervised finetuning paradigm, further advancing code translation performance without relying on any auxiliary datasets.

\paragraph{Reinforcement Learning from Human Feedback.}
RLHF has significantly improved the performance of downstream tasks by aligning models with human preferences. 
While the reward-based methods like proximal policy optimization \cite{schulman2017proximal} and group relative policy optimization \cite{shao2024deepseekmath} demonstrate success, they suffer from increased memory overhead and substantial computational burdens due to their requirement for a large number of samples during each policy update cycle. 

Reward-free methods address these limitations from a different perspective.
Direct preference optimization \cite{rafailov2024direct} eliminates the need for an explicit reward model by implicitly estimating reward signals from preference data. Identity preference optimization \cite{azar2024general} introduces a general non-decreasing bounded function to mitigate overfitting, while simple preference optimization \cite{meng2025simpo} further simplifies the process by removing the reference model altogether. 
Additionally, some studies \cite{park2024disentangling,meng2025simpo} explore the regularization terms as a margin to improve reward modeling. 
Despite these advancements, directly applying these techniques to code translation remains suboptimal due to inadequate syntax representation and incomplete semantic modeling. 
To address this challenge, \methodname enhances preference optimization by jointly incorporating syntax and semantics, leading to improved accuracy and robustness in code translation.

\section{Methodology}

\begin{figure*}
\centering
\includegraphics[width=\linewidth]{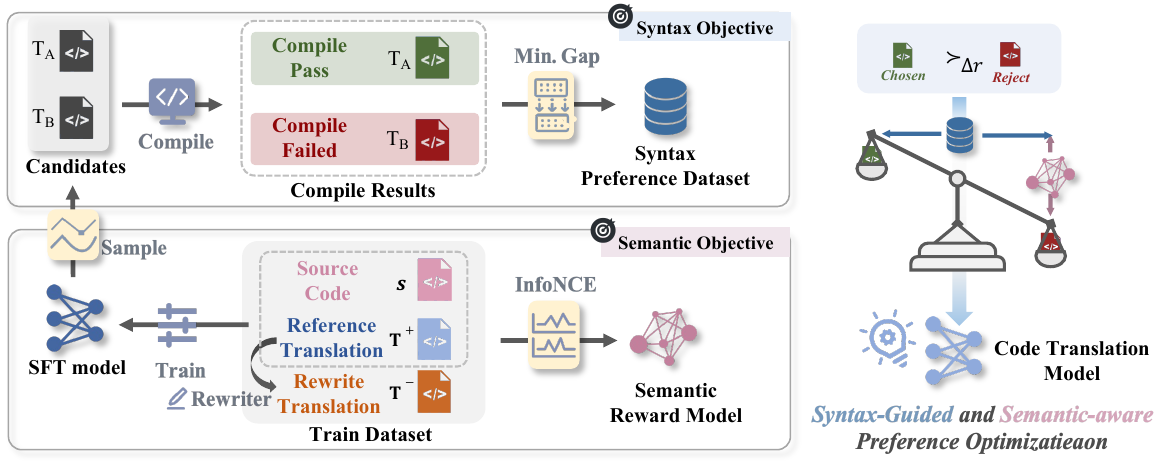}
\caption{Overview of our \methodname.}
\label{fig:approach}
\end{figure*}

\subsection{Reward Model Training}
\label{sec:rewardmodel}
The overall reward comprises both syntactic and semantic components. 
The syntactic reward is readily accessible, as it can be directly derived from the compiler feedback. 
This binary and rule-based signal indicates whether the translation code adheres to the target language’s grammar, following standard practice in prior work \cite{shojaee2023executionbased}.

In contrast, designing an effective semantic reward presents a more challenging problem, as code semantics encompass not only functional equivalence, but also logical fidelity to the source code’s intent. Prior studies have approached this from different perspectives.
Execution-based reward like CodeRL \cite{le2022coderl} utilizes discrete, rule-based heuristics to evaluate semantic correctness, primarily relying on unit tests. 
However, test cases are often scarce or even unavailable in function- or program-level translation training data, resulting in unstable execution-based reward signals. 
Furthermore, such reward is inherently binary and sparse, making it difficult to quantify the proximity of an incorrect translation to the correct logic. 
Reference-based reward like PPOCoder \cite{shojaee2023executionbased} employs reward signals derived from static program representations, such as abstract syntax trees and dataflow graphs, which often depend on comparisons with ground-truth reference implementations. 
The reliance on reference translations restricts their applicability in real-world scenarios, where such references are often incomplete or entirely absent. 
Moreover, minor code variations, such as variable renaming, can lead to significant runtime or functional errors, which are often not adequately penalized by textual metrics.

\paragraph{Semantic Reward Model.}
To address these issues, we propose to assess semantic correctness by measuring semantic distance within a high-dimensional latent space. 
Formally, our objective is to learn a mapping function $f: \mathcal{X} \cup \mathcal{Y} \to \mathcal{Z}$ such that the geometric proximity in $\mathcal{Z}$ reflects the functional equivalence between the source code $x$ and the target code $y$.
To shape this latent space, we construct a source-anchored triplet $\mathcal{T} = (x, y^+, \mathcal{Y}^-)$ for each training instance. 
We designate the source code $x$ as the fixed anchor point in the latent space. It serves as the semantic oracle which all translation candidates are measured. 
The reference translation $y^+$ is defined as the positive sample and $\mathcal{Y}^- = \{y^-_1, \dots, y^-_K\}$ is semantically divergent negatives.
The optimization goal is to pull the embedding $f(y^+)$ into the immediate $\epsilon$-neighborhood of the anchor $f(x)$, while explicitly repelling the hard negatives in $\mathcal{Y}^-$ beyond this neighborhood boundary, thereby establishing a robust semantic margin against subtle deviations.

\paragraph{Training Objective.}
Our semantic reward model adopts a dual-encoder architecture, which contains a source code encoder and a target code encoder, sharing the same parameters. 
To effectively learn semantic similarity for cross-lingual code representation, we use the InfoNCE loss \cite{oord2018representation} as the training objective:
\begin{align}
\resizebox{\columnwidth}{!}{$
\mathcal{L}_f = - \log \frac{\exp \left( \cos(\frac{f(\mathbf{x}), f(\mathbf{y}^+)}{\tau} \right)}{\exp \left( \cos(\frac{f(\mathbf{x}), f(\mathbf{y}^+)}{\tau } \right) + \sum\limits_{i=1}^{K} \exp \left( \cos(\frac{f(\mathbf{x}), f(\mathbf{y}^-_i)} {\tau} \right)},  
$}
\end{align}
where $\mathbf{x}$ denotes the source code, $\mathbf{y}^+$ and $\mathbf{y}^-_i$ denotes reference translation and negative translations respectively.
 $f(\cdot)$ is the encoder that maps code snippets to semantic representations, $\cos(\cdot, \cdot)$ denotes cosine similarity, and $\tau$ is the temperature parameter.

\paragraph{Dataset for Training.}
The positive samples come from the reference code in the supervised dataset. To construct negative samples for training the semantic reward model, we employ an LLM as a perturbation generator. Specifically, the LLM rewrites the reference code by introducing subtle perturbations that alter the intended semantics.
These perturbed variants, which often remain syntactically valid but semantically flawed, making them ideal negative examples. Together with the original reference translations, they form the training dataset for learning cross-lingual semantic alignment.

\paragraph{Reward Score Function.}
Based on the semantic model, we encode both the source and candidate target code into embeddings within a shared vector space.
For a set of candidate translations $y = \{y_1, \dots, y_n\}$ in the target language, we compute the cosine similarity between each candidate $y_i$ and the source code $x$. 
The similarity scores are transformed into logit values and standardized via z-score normalization to produce the final semantic reward:
\begin{align}
\begin{split}
    r_s(y_i) &= \frac{s_i - \mu(s_1, \dots, s_n)}{\sigma(s_1, \dots, s_n)},\\
    s_i &= \log\frac{\cos(f(x), f(y_i))}{1 - \cos(f(x), f(y_i))},
\end{split}
\end{align}
where $\mu$ and $\sigma$ represent the mean and standard deviation of the scores ${s_1, \dots, s_n}$, respectively. 
This function enables list-wise scoring by capturing the semantic similarity between the source code and each candidate translation. 

\subsection{Preference Optimization}

\paragraph{Problem Formulation.}
We formulate code translation as a multi-objective optimization problem over syntax and semantics, and apply a linear scalarization strategy to integrate the multiple objectives into a unified optimization goal.
We denote the syntax objective as $g$, the semantics objective as $s$, and the preference dataset for code translation as
$\mathcal{D} = \{ \mathcal{D}_g, \mathcal{D}_s \}$,
where $\mathcal{D}_g$ and $\mathcal{D}_s$ denote the syntactic preference dataset and semantic preference dataset, respectively.
Given the dataset $\mathcal{D}$, we define the oracle reward model as a weighted combination of syntax and semantic rewards:
\begin{equation}
    \mathbf{r}^*(\mathbf{x}, \mathbf{y}) = w \cdot r^*_g(\mathbf{x}, \mathbf{y}) + (1 - w) \cdot r^*_s(\mathbf{x}, \mathbf{y}),
\end{equation}
where $r^*_g(\mathbf{x}, \mathbf{y})$ and $r^*_s(\mathbf{x}, \mathbf{y})$ evaluate the syntax and semantic rewards of the translated target code, respectively.
$w \in [0,1]$ is a weight parameter to balance syntactic and semantic preferences.
In contrast to the classical Pareto optimization, which seeks to approximate the Pareto front of multiple objectives, we focus on optimizing the code translation model under a specific preference weighting $w=0.5$. 
This choice represents an equal prioritization of syntactic and semantic objectives.

This leads to the following training objective for the translation model $\pi_\theta$:
\begin{equation}
\label{eq3}
    \max_{\pi_\theta} \; \mathbb{E}_{\mathbf{x}, \mathbf{y} \sim \pi(\cdot\,|\,\mathbf{x})} \left[ \mathbf{r}^*(\mathbf{x}, \mathbf{y}) - \beta \log \frac{\pi_\theta(\mathbf{y}\,|\,\mathbf{x})}{\pi_{\text{sft}}(\mathbf{y}\,|\,\mathbf{x})} \right],
\end{equation}
where $\pi_{\text{sft}}$ is the supervised finetuned model serving as the reference policy, $\beta$ is a regularization coefficient controlling deviation from the reference model.

While previous works \cite{shojaee2023executionbased} have leveraged RL for reward-based finetuning, such approaches tend to be unstable and resource-intensive.
We propose an RL-free variant of DPO, designed to integrate syntax and semantics into a unified training process efficiently.

\paragraph{Derivation of \methodname.}
The optimal policy under the RLHF objective (Eq.~\eqref{eq3}) takes the form:
\begin{align}
\label{eq5}
    \pi^*_\theta(y\,|\,x) = \frac{1}{Z(x)} \pi_\text{sft}\left(y\,|\,x\right) \exp\left(\frac{\mathbf{r}^*(x, y)}{\beta}\right),
\end{align}
where the partition function is defined as: 
$Z(x) = \sum_y \pi_\text{sft}(y\,|\,x) \exp\left( \frac{1}{\beta} \left( w \cdot r^*_g(x, y) + (1 - w) \cdot r^*_s(x, y) \right) \right)$.

Let $\mathcal{D}_k = \{(x, y_w, y_l)\}$ denote the pairwise preference dataset under objective $k \in \{g, s\}$, where $y_w \succ y_l$ indicates that $y_w$ is preferred over $y_l$ for input $x$. The pairwise preference likelihood under the oracle reward model is given by:
\begin{equation}
\label{eq6}
    p_{\mathcal{D}_k}(y_w \succ y_l \mid x) = \sigma\left( \mathbf{r}^*(x, y_w) - \mathbf{r}^*(x, y_l) \right),
\end{equation}
where $\sigma(\cdot)$ denotes the sigmoid function.

Substituting Eq.~\eqref{eq5} into Eq.~\eqref{eq6} and approximating the ground-truth reward $r^*_{-k}(x, y)$ from the complementary objective $-k$ using a learned reward model $r_{-k}(x, y)$, we derive the final training objective of \methodname:
\begin{align}
\label{eq9}
&\mathcal{L}_{\text{\methodname}} = 
    - \,\mathbb{E}_{(x, y_w, y_l) \sim \mathcal{D}_k}\bigg[
    \log \sigma \bigg( \frac{\beta}{w} \Big( \log \frac{\pi_\theta(y_w \,|\, x)}{\pi_{\text{sft}}(y_w \,|\, x)} \notag\\
    &\ - \log \frac{\pi_\theta(y_l \,|\, x)}{\pi_{\text{sft}}(y_l \,|\, x)}\Big)
    - \frac{1-w}{w}\Big( r_{-k}(x, y_w) - r_{-k}(x, y_l) \Big)\bigg) 
    \bigg],
\end{align}
where $-k$ denotes the objective complementary to $k$ (i.e., if $k = g$, then $-k = s$, and vice versa). 

This formulation reveals the core mechanism of \methodname: by leveraging preference data from one objective (captured in $\mathcal{D}_k$) and integrating the explicit reward difference from the other ($r_{-k}$) as a rectification term, we realize the simultaneous modeling of both syntactic and semantic objectives within a unified optimization process.

\paragraph{Preference Dataset Construction.}
Constructing a high-quality preference dataset is essential for effective model training. 
Syntactic feedback is generally more stable and verifiable, as grammar correctness can be deterministically checked via compilation. 
Such feedback serves as an oracle-like signal that enables the reliable classification of candidate translations into preferred and less preferred sets.
In contrast, semantic feedback is inherently more ambiguous. 
Although extended unit tests and semantic reward models offer approximate semantic judgments, determining the absolute correctness of a translation remains an undecidable problem.
Given this discrepancy in reliability, we build our preference dataset primarily based on syntactic correctness, where preferences between candidate translations are derived from deterministic compilation results. 

\begin{table*}
    \centering
    \resizebox{.85\textwidth}{!}{
    \begin{tabular}{l|cccccc|cccccc}
        \toprule
        \multirow{2}{*}{Methods} & \multicolumn{6}{c|}{TransCoder-Test} & \multicolumn{6}{c}{HumanEval-X} \\ 
        \cmidrule(lr){2-7} \cmidrule(lr){8-13} & C$\rightarrow$J & C$\rightarrow$P & J$\rightarrow$C & J$\rightarrow$P & P$\rightarrow$C & P$\rightarrow$J & C$\rightarrow$J & C$\rightarrow$P & J$\rightarrow$C & J$\rightarrow$P & P$\rightarrow$C & P$\rightarrow$J \\
        \midrule
        TransCoder & 66.59 & 43.75 & 80.51 & 49.57 & 34.05 & 36.51 & 48.17 & 38.41 & 38.41 & 42.68 & 20.73 & 26.83 \\
        TransCoder-ST & 68.26 & 65.95 & \textbf{82.23} & 72.41 & 57.38 & 56.43 & 46.34 & 40.85 & 40.85 & 64.63 & 21.95 & 23.17 \\
        CodeT5-SFT & 73.65 & 70.91 & 80.73 & 72.41 & 65.95 & 66.39 & 47.56 & 43.90 & 45.73 & 65.24 & 27.44 & 28.66 \\
        PPOCoder & 72.20 & 57.11 & 65.74 & 52.15 & 38.97 & 51.24 & 42.68 & 40.85 & 42.68 & 59.76 & 26.83 & \textbf{29.27} \\
        \methodname (CodeT5) & \textbf{75.73} & \textbf{74.57} & \textbf{82.23} & \textbf{75.43} & \textbf{67.66} & \textbf{68.26} & \textbf{50.00} & \textbf{45.73} & \textbf{49.39} & \textbf{69.51} & \textbf{29.27} & \textbf{29.27} \\

        \midrule
        CodeLlama-7B & 62.86 & 68.32 & 56.71 & 68.97 & 65.52 & 49.37 & 75.61 & 62.80 & 50.00 & 78.66 & 42.68 & 51.83 \\
        CodeLlama-7B-SFT & 80.91 & 78.23 & 79.01 & 81.47 & 80.73 & 77.39 & 82.93 & 68.29 & 59.15 & 82.32 & 50.00 & 56.71 \\
        \methodname (CodeLlama-7B) & \textbf{86.51} & \textbf{81.90} & \textbf{82.87} & \textbf{83.83} & \textbf{82.23} & \textbf{79.46} & \textbf{84.15} & \textbf{74.39} & \textbf{65.85} & \textbf{83.54} & \textbf{50.61} & \textbf{59.15} \\

        \midrule
        Qwen2.5-Coder-7B & 71.37 & 78.23 & 42.40 & 79.31 & 60.38 & 27.19 & 77.44 & 74.39 & 70.12 & 81.71 & 56.71 & 68.29 \\
        Qwen2.5-Coder-7B-SFT & 87.55 & 82.97 & 91.65 & 87.07 & 86.93 & 84.85 & 84.76 & 77.44 & 72.56 & 86.59 &62.80 & 70.12 \\
        \methodname (Qwen2.5-Coder-7B) & \textbf{90.87} & \textbf{87.93} & \textbf{93.36} & \textbf{90.95} & \textbf{89.29} & \textbf{85.89} & \textbf{87.19} & \textbf{82.31} & \textbf{73.78} & \textbf{87.19} & \textbf{64.02} & \textbf{71.95} \\
        \bottomrule
    \end{tabular}}
    \caption{CA@1 scores on the TransCoder-Test and HumanEval-X datasets.
    ``C'', ``J'', and ``P'' denote C++, Java, and Python, respectively.
    The best scores are marked in \textbf{bold}.}
    \label{tab:transcoder_main}
\end{table*}

As shown in Figure~\ref{fig:approach}, we begin with a supervised dataset comprising aligned source and target code pairs. A base code LLM is finetuned on this dataset to improve its code translation capabilities. 
For a source code snippet $x$, we generate a set of candidate translations $\{ y_1, y_2, \dots, y_n \}$ using the finetuned model. 
Each candidate $y_i$ is passed through a compiler to obtain a binary signal indicating whether the code compiles successfully.
Based on the compilation results, the candidates are partitioned into a pass set and a fail set, from which we construct a preference dataset. 
Instead of selecting random pairs, we employ \texttt{git-diff}\footnote{\url{https://git-scm.com/docs/git-diff}} to identify the closest matching pair $(y_w,y_l)$, ensuring that low-gap pairs capture fine-grained preferences, where $y_w$ comes from the pass set and $y_l$ comes from the failed set.
We then incorporate scores from the semantic reward model as soft guidance signals. Specifically, the semantic scores of $y_w$ and $y_l$ are used during optimization to enrich the syntactic preference with semantic discrimination.

This strategy leverages the determinism of syntactic preference labels while still benefiting from the fine-grained semantic discrimination offered by the semantic reward model.
Such integration facilitates the simultaneous optimization of syntactic accuracy and semantic alignment.

\section{Experiments and Results}

\subsection{Experiment Settings}

\paragraph{Dataset Construction.}
Our supervised finetuning training set and preference dataset construction are based on the XLCoST dataset \cite{zhu2022xlcost}, which contains parallel snippet-level and program-level code for commonly used programming languages, with each sample accompanied by an example test case to verify functionality.
We collect the parallel program-level data from the training and validation splits. 
The supervised finetuning training set consists of 6,884 Java$\leftrightarrow$C++, 6,419 C++$\leftrightarrow$Python, and 7,278 Java$\leftrightarrow$Python samples, with a validation set of 346, 323, and 376 samples for each pair, respectively.
After finetuning, we generate 10 candidate responses for each input using a sampling temperature of 0.9.

For the semantic model training dataset, we employ the Qwen3-8B model to rewrite the reference code of XLCoST, thereby generating corresponding negative samples. 
Because the inference is conducted on candidates sampled from the supervised finetuning model at the semantic model's test time, rather than on the reference data itself, this avoids data leakage and ensures benchmarks remain entirely unseen by the semantic model.

\paragraph{Benchmarks.}
To assess the performance of our \methodname, we employ two benchmark datasets:
\begin{itemize}
    \item  \textbf{TransCoder-Test} \cite{roziere2020transcoder} is a standard benchmark for unsupervised code translation, with tasks in Java, C++, and Python (482, 467, and 464 tasks, respectively) and an average of 10 test cases per task.

    \item  \textbf{HumanEval-X} \cite{zheng2023codegeex} is extended from the HumanEval benchmark \cite{chen2021codex} and designed for cross-lingual code correctness evaluation using automated test cases. It includes 164 tasks, with 6.9 test cases on average.
\end{itemize}

\paragraph{Evaluation Metrics.}
Computational Accuracy at top-K (CA@K) measures the proportion of tasks where at least one of the top-K translations generated by the model passes all test cases. 
We use CA@1, showing the model's effectiveness in generating functionally equivalent code to the reference.

\begin{table*}
    \centering
    \resizebox{.85\textwidth}{!}{
    \begin{tabular}{l|cccccc|cccccc}
        \toprule
        \multirow{2}{*}{Methods} & \multicolumn{6}{c|}{TransCoder-Test} & \multicolumn{6}{c}{HumanEval-X} \\ 
        \cmidrule(lr){2-7} \cmidrule(lr){8-13} & C$\rightarrow$J & C$\rightarrow$P & J$\rightarrow$C & J$\rightarrow$P & P$\rightarrow$C & P$\rightarrow$J & C$\rightarrow$J & C$\rightarrow$P & J$\rightarrow$C & J$\rightarrow$P & P$\rightarrow$C & P$\rightarrow$J \\
        
        \midrule
        \methodname (CodeLlama-7B) & \textbf{86.51} & \textbf{81.90} & \textbf{82.87} & \textbf{83.83} & \textbf{82.23} & \textbf{79.46} & \textbf{84.15} & \textbf{74.39} & \textbf{65.85} & \textbf{83.54} & \textbf{50.61} & \textbf{59.15} \\
        \quad w/o semantic & 82.31 & 81.47 & 80.08 & 81.90 & 81.80 & 79.25 & 82.78 & 73.78 & 60.98 & 83.54 & 50.00 & 58.54\\
        \quad w/o syntax & 82.16 & 79.09 & 80.30 & 82.33 & \textbf{82.23} & \textbf{79.46} & 82.32 & 65.85 & 60.98 & 80.49 & 48.78 & \textbf{59.15}\\
        IPO (CodeLlama-7B) & 83.82 & 75.43 & 82.65 & 83.62 & 80.09 & 44.19 & 40.24 & 66.46 & 63.41 & 81.10 & 48.78 & 23.78\\
        SimPO (CodeLlama-7B) & 83.40 & 73.71 & 81.58 & 83.40 & 80.30 & 77.80 & 41.46 & 65.85 & 63.41 & 81.70 & 48.78 &26.83 \\
        \midrule
        \methodname (Qwen2.5-Coder-7B) & \textbf{90.87} & \textbf{87.93} & \textbf{93.36} & \textbf{90.95} & \textbf{89.29} & \textbf{85.89} & \textbf{87.19} & \textbf{82.31} & \textbf{73.78} & \textbf{87.19} & \textbf{64.02} & \textbf{71.95} \\
        \quad w/o semantic & 89.63 & 86.21 & 92.51 & 87.93 & 89.08 & 85.06 & 85.98 & 78.66 & 73.17 & 86.59 & 62.20 & \textbf{71.95} \\
        \quad w/o syntax & 89.00 & 83.19 & 93.15 & 86.85 & 86.51 & 85.06 & 85.36 & 75.00 & 71.95 & 86.59 & 63.41 & 70.73\\
        IPO (Qwen2.5-Coder-7B) & 89.00 & 87.07 & 92.93 & 89.87 & 88.44 & 85.27 & 44.51 & 79.87 & \textbf{73.78} & 85.98 & 60.98 & 34.14\\
        SimPO (Qwen2.5-Coder-7B) & 90.04 & 87.28 & \textbf{93.36} & 89.01 & 87.79 & 85.27 & 44.51 & 79.87 & 73.17 & 85.98 & 62.20 & 34.75 \\
        \bottomrule
    \end{tabular}}
    \caption{CA@1 scores of ablation study and alternative methods on the TransCoder-Test and HumanEval-X datasets.}
    \label{tab:transcoder_ablation}
\end{table*}

\paragraph{Baseline Methods.}
We compare \methodname to two categories of code translation methods:
\begin{itemize}
\item \textbf{Unsupervised translation} does not require parallel corpora for training. Instead, it leverages self-supervised learning techniques:
(1) \textbf{TransCoder} \cite{roziere2020transcoder} is an unsupervised model with approximately 100M parameters that leverages masked language modeling, denoising autoencoding, and back translation as its core pre-training tasks.
(2) \textbf{TransCoder-ST} \cite{roziere2022leveraging} extends TransCoder by incorporating automated unit tests to reduce noise in back translation, thereby improving model performance.

\item \textbf{Supervised translation} relies on parallel corpora with aligned source and target code pairs to learn a transformation between different programming languages:
(1) \textbf{CodeT5-SFT} is a supervised finetuned variant of CodeT5 \cite{wang2021codet5}.
We choose CodeT5 as the base model due to its comparable scale (770M) and architecture to TransCoder and TransCoder-ST, ensuring a balanced comparison.
(2) \textbf{PPOCoder} \cite{shojaee2023executionbased} is built upon CodeT5-SFT with proximal policy optimization (PPO). 
It incorporates compiler feedback and both syntactic and semantic match scores as reward signals to improve translation quality.
(3) \textbf{CodeLlama-7B-SFT} is a supervised finetuned variant of CodeLlama-7B \cite{roziere2023codellama}, which serves as a representative foundational model for code-related tasks.
(4) \textbf{Qwen2.5-Coder-7B-SFT} is a supervised finetuned variant of Qwen2.5-Coder-7B \cite{hui2024qwen2}.
\end{itemize}

\paragraph{Implementation.}
For supervised finetuning, CodeT5 uses a learning rate of 5e-5 and a batch size of 128, while CodeLlama-7B and Qwen2.5-Coder-7B use 3e-5 and 32, respectively, with gradient accumulation for memory constraints. 
For the semantic reward model, we employ Qwen3-Embedding 0.6B as the base model. The training is conducted with a learning rate of 6e-6 and a batch size of 32 for 5 epochs.
For preference optimization, we set the learning rate to 5e-7 for CodeT5-SFT, and 5e-5 for both CodeLlama-SFT and Qwen2.5-Coder-7B.
For CodeLlama-7B and Qwen2.5-Coder-7B, we apply LoRA \cite{hu2022lora} for all finetuning processes, including both supervised finetuning and preference optimization phases.

All experiments are conducted on one NVIDIA RTX A800 GPU and Ubuntu 20.04 LTS.
See our source code\footnote{\url{https://github.com/nju-websoft/CTO}} for other implementation details.

\subsection{Code Translation Results}

Table~\ref{tab:transcoder_main} shows the experimental results on the TransCoder-Test and HumanEval-X datasets. 
Our \methodname consistently outperforms existing methods across six translation tasks. 

Compared with the unsupervised code translation methods, \methodname (CodeT5) surpasses both TransCoder and TransCoder-ST across all evaluated scenarios.
This performance gap underscores the efficacy of supervised finetuning on parallel corpora, which facilitates the direct acquisition of cross-lingual mappings and ensures higher translation reliability. 
In contrast, unsupervised methods rely on back translation, which can introduce noise and misalignment.
For PPOCoder, we follow its reward function and parameter settings in \cite{shojaee2023executionbased} and apply PPO on our finetuned CodeT5 model. 
Contrary to our expectations, PPOCoder exhibits performance degradation in most cases.
This implies that PPOCoder does not always benefit from its reward function and may even introduce instability to the finetuned model. 

Since the preference dataset is derived from supervised finetuned models, its quality depends on the model's ability to generate valid compilable samples. 
However, CodeT5-SFT generates fewer valid samples in translation tasks from Python, making preference data collection more difficult. 
This observation is consistent with previous studies \cite{feng2024towards}, which have noted a strong correlation between the performance of supervised finetuned models and the effectiveness of preference optimization.

To assess the scalability of \methodname on larger code models, we conduct experiments using decoder-only LLMs, CodeLlama-7B and Qwen2.5-Coder-7B.
Across six language pairs and two benchmarks, \methodname consistently outperforms their supervised finetuned counterparts (CodeLlama-7B-SFT and Qwen2.5-Coder-7B-SFT).
These improvements validate the generalization of \methodname across different model architectures and parameter sizes. 

\subsection{Ablation Study and Alternative Methods}

\begin{table*}
    \centering
    \resizebox{.85\textwidth}{!}{
    \begin{tabular}{l|cccccc|cccccc}
    \toprule
        \multirow{2}{*}{Methods} & \multicolumn{6}{c|}{TransCoder-Test} & \multicolumn{6}{c}{HumanEval-X} \\ 
        \cmidrule(lr){2-7} \cmidrule(lr){8-13} & C$\rightarrow$J & C$\rightarrow$P & J$\rightarrow$C & J$\rightarrow$P & P$\rightarrow$C & P$\rightarrow$J & C$\rightarrow$J & C$\rightarrow$P & J$\rightarrow$C & J$\rightarrow$P & P$\rightarrow$C & P$\rightarrow$J \\
        \midrule
        \methodname (CodeLlama-7B) & \textbf{86.51} & \textbf{81.90} & \textbf{82.87} & \textbf{83.83} & \textbf{82.23} & \textbf{79.46} & \textbf{84.15} & 74.39 & \textbf{65.85} & \textbf{83.54} & \textbf{50.61} & \textbf{59.15} \\
        \quad w/ CodeBLEU & 83.20 & 81.47 & 82.01 & 82.76 & \textbf{82.23} & 73.86 & 82.93 & \textbf{75.61} & 62.80 & 81.71 & \textbf{50.61} & 57.93 \\
        \quad w/ CodeBERTScore & 83.61 & 81.25 & 82.23 & 82.54 & 82.01 & 73.44 & 82.93 & \textbf{75.61} & 62.80 & 82.32 & \textbf{50.61} & 57.32 \\
        \midrule
        \methodname (Qwen2.5-Coder-7B) & \textbf{90.87} & \textbf{87.93} & \textbf{93.36} & \textbf{90.95} & \textbf{89.29} & \textbf{85.89} & \textbf{87.19} & \textbf{82.31} & \textbf{73.78} & \textbf{87.19} & \textbf{64.02} & 71.95 \\
        \quad w/ CodeBLEU & 88.80 & 87.28 & 92.72 & 88.58 & 88.44 & 84.65 & 86.59 & 79.27 & 73.17 & 86.59 & 62.80 & 72.56 \\
        \quad w/ CodeBERTScore & 89.63 & 87.07 & 92.29 & 88.58 & 88.87 & 85.27 & 86.59 & 79.88 & 73.17 & 85.98 & 63.41 & \textbf{73.17} \\
        \bottomrule
    \end{tabular}}
    \caption{Comparison of CA@1 scores between reference-based metrics and the semantic reward model.}
    \label{tab:semantic_ablation}
\end{table*}

To investigate the effectiveness of multi-objective preference optimization in our proposed method, we conduct an ablation study by systematically removing key components and analyzing their impact. 
To measure the contribution of multi-objective preference optimization, we design two variants:
\begin{itemize}
    \item \textbf{\methodname w/o semantic} removes the semantic preference from the preference optimization stage, meaning the model is optimized solely based on the syntax preference data. 
    Consequently, \methodname degenerates into DPO, which helps isolate the impact of semantic reward.

    \item \textbf{\methodname w/o syntax} removes the syntax preference from the preference optimization stage, i.e., the model is optimized solely based on the semantic preference data. In this setting, \methodname degenerates to vanilla DPO, where the reference translation is treated as the chosen sample and the rejected samples are standard outputs that do not match the reference.
    
\end{itemize}

As shown in Table~\ref{tab:transcoder_ablation}, the results clearly indicate that each preference contributes significantly to the final performance.  
Notably, in the w/o syntax variant, we rely on the example unit test shipped with the original dataset as auxiliary signals to guide optimization.
But the performance still falls short compared to \methodname.
These results confirm that multi-objective preference plays a crucial role in maximizing performance.

Additionally, to further validate the effectiveness of our preference optimization strategy, we compare our method with two typical reward-free preference optimization techniques: identity preference optimization (IPO) \cite{azar2024general} and simple preference optimization (SimPO) \cite{meng2025simpo}. 
These baselines perform preference optimization on supervised finetuned models, enabling us to assess whether \methodname provides a tangible advantage over existing techniques. 
As also presented in Table~\ref{tab:transcoder_ablation}, \methodname achieves higher performance across most translation tasks on both TransCoder-Test and HumanEval-X benchmarks. 
While IPO and SimPO fail to match the performance of our method in certain scenarios, they even cause performance degradation when applied to supervised finetuned models. 

Overall, these results demonstrate that our \methodname effectively integrates syntactic and semantic modeling within a preference optimization framework, leading to superior performance in code translation tasks.
These findings reinforce the importance of multi-objective preference optimization as a robust strategy for enhancing code translation models.

\subsection{Reward Model Evaluation}

\begin{figure}
    \centering
    \includegraphics[width=\linewidth]{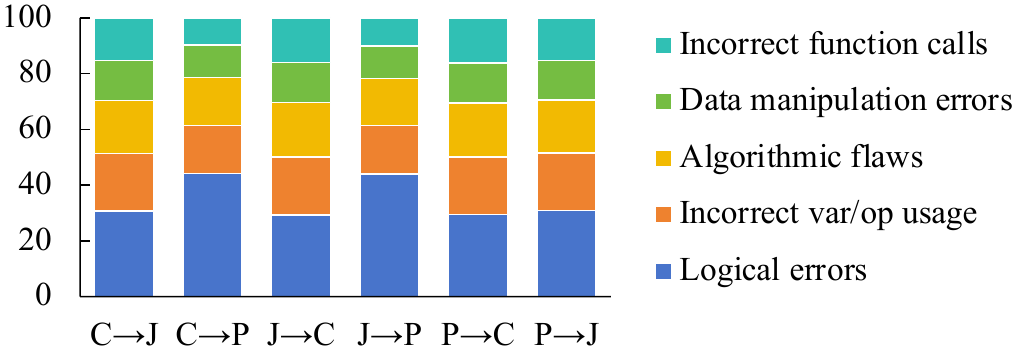}
    \caption{Distribution of negative sample types.}
    \label{fig:errors}
\end{figure}

\paragraph{Negative Sample Types.}
To evaluate the quality and diversity of our semantic reward model's training data, we analyze the distribution of various negative sample types, following the semantic error classification outlined in \cite{pan2024lost}. 
As shown in Figure~\ref{fig:errors}, the categorization reveals that logical errors constitute the largest portion of the dataset.
The extensive coverage of error categories, coupled with their relatively balanced representation, prevents the model from being biased toward specific patterns and ensures robust detection of various semantic discrepancies.

\begin{figure}
    \centering
    \includegraphics[width=\linewidth]{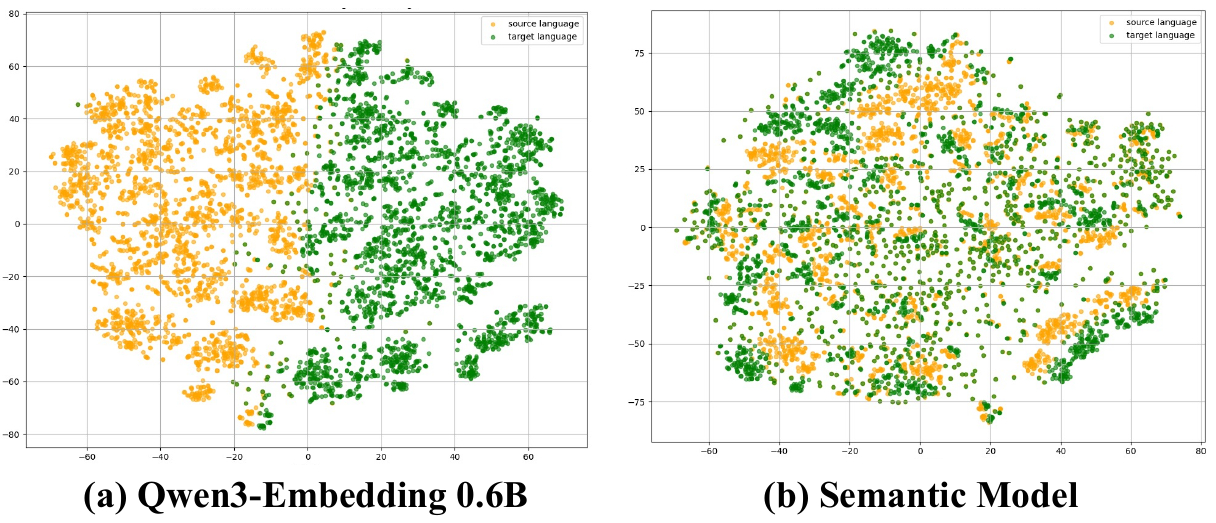}
    \caption{t-SNE representation of source language code (yellow) and target language code (green).}
    \label{fig:latent}
\end{figure}

\paragraph{Latent Space Visualization.}
To assess the effectiveness of our trained reward model, we conduct an evaluation of our semantic reward model and visualize the source and target code embeddings derived from the reward model in a shared vector space. Figure~\ref{fig:latent} shows that embeddings from different programming languages form coherent clusters instead of being strictly segregated by language. This verifies the model's ability to fuse cross-lingual code semantics into a unified embedding space, suggesting that it captures meaningful alignment between source and target language structures.

\paragraph{Effectiveness of the Semantic Reward Model.}
Lastly, we compare the performance of our semantic reward model and previous reference-based metrics such as CodeBLEU \cite{Ren2020codebleu} and CodeBERTScore \cite{zhou2023codebertscore} when applied to \methodname. 
As presented in Table~\ref{tab:semantic_ablation}, our semantic reward model consistently outperforms both CodeBLEU and CodeBERTScore across most scenarios. 
This performance gap highlights the advantage of leveraging our learned semantic model over relying solely on reference-based metrics as the semantic reward. 

Overall, these results confirm the cross-lingual semantic coherence captured by our semantic reward model and its superior effectiveness over prior reference-based metrics.

\section{Conclusion}
\label{sec:concl}

In this paper, we present \methodname to improve code translation by syntax-guided and semantic-aware preference optimization. 
We formulate code translation as a multi-objective preference optimization problem. 
By leveraging compiler feedback and semantic reward modeling, \methodname effectively refines translation quality, ensuring both syntactic correctness and semantic alignment.
Experimental results across multiple benchmarks demonstrate that \methodname significantly outperforms state-of-the-art methods in code translation accuracy and robustness. 
The proposed preference optimization approach is also generalizable, making it adaptable to various programming languages and model architectures.
In future work, we will extend \methodname to support a broader range of programming languages and explore its effectiveness in other software migration scenarios.

\section*{Acknowledgments}

This work was supported by the ``111 Center'' (No. B26023), the Fundamental and Interdisciplinary Disciplines Breakthrough Plan of the Ministry of Education of China (No. JYB2025XDXM118), and the Cooperation Fund of Huawei Cooperation Project (No. TC20230202021-2024-12).

\bibliographystyle{named}
\bibliography{ijcai26}

\end{document}